\documentclass[10pt,twocolumn,letterpaper]{article}

\usepackage{cvpr}              
\usepackage{algorithm}
\usepackage{ulem}
\usepackage{algorithmic}

\definecolor{cvprblue}{rgb}{0.21,0.49,0.74}
\usepackage[pagebackref,breaklinks,colorlinks,allcolors=cvprblue]{hyperref}

\def\confName{CVPR}
\def\confYear{2025}

\title{PhyCAGE: Physically Plausible Compositional 3D Asset Generation from a Single Image}

\author{Han Yan\textsuperscript{1}\footnotemark[1] \quad Mingrui Zhang\textsuperscript{2} \quad Yang Li\textsuperscript{2} \quad Chao Ma\textsuperscript{1} \quad Pan Ji\textsuperscript{2} \\
\textsuperscript{1} MoE Key Lab of Artificial, AI Institute, Shanghai Jiao Tong University\\
\textsuperscript{2} Tencent XR Vision Labs\\
\url{https://wolfball.github.io/phycage}
}

\begin{document}

\twocolumn[{%
\renewcommand\twocolumn[1][]{#1}%
\maketitle
\begin{center}
    \centering
    \captionsetup{type=figure}
    \includegraphics[width=.74\textwidth]{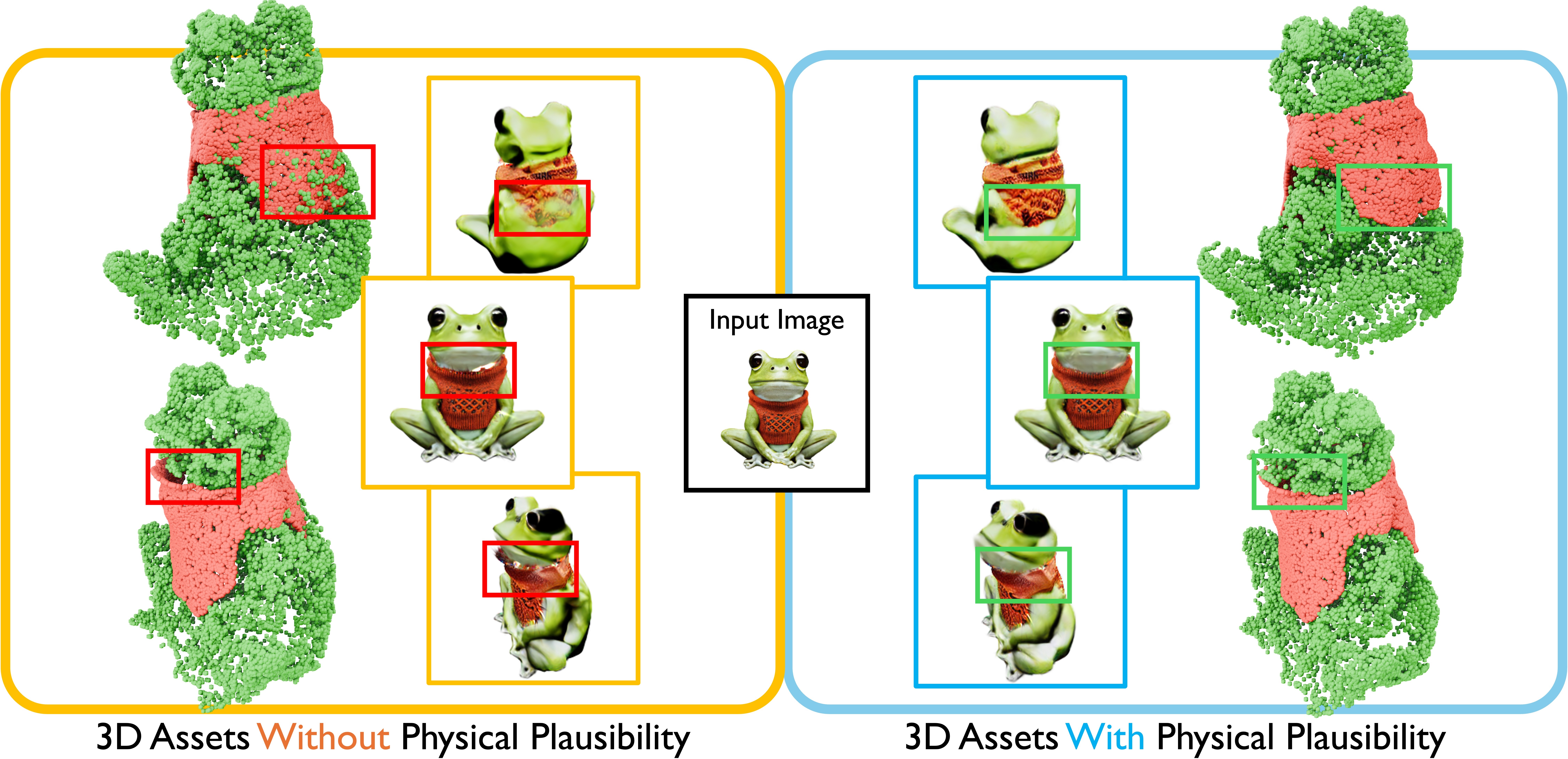}
    \captionof{figure}{PhyCAGE can generate compositional 3D assets with interactive objects in a physically plausible manner. The generated 3D Gaussian Splatting shows better visual performance and physical plausibility under Material Point Method (MPM) simulation.}
\end{center}%
}]

\footnotetext[1]{~Work done during internship at Tencent XR Vision Labs.}

\begin{abstract}

We present \textbf{PhyCAGE}, the first approach for \textbf{Phy}sically plausible \textbf{C}ompositional 3D \textbf{A}sset \textbf{Ge}neration from a single Image.
Given an input image, we first generate consistent multi-view images for components of the assets. These images are then fitted with 3D Gaussian Splatting representations.
To ensure that the Gaussians representing objects are physically compatible with each other, we introduce a Physical Simulation-Enhanced Score Distillation Sampling (PSE-SDS) technique to further optimize the positions of the Gaussians.
It is achieved by setting the gradient of the SDS loss as the initial velocity of the physical simulation, allowing the simulator to act as a physics-guided optimizer that progressively corrects the Gaussians' positions to a physically compatible state.
Experimental results demonstrate that the proposed method can generate physically plausible compositional 3D assets given a single image.

\end{abstract}    
\section{Introduction}
\label{sec:intro}

Generating 3D shapes conditioned on 2D image input lies at the core of many applications, such as virtual reality (VR), augmented reality (AR), video gaming, and robotics.
Recently, this field has seen remarkable progress, thanks to advancements in AI techniques, including transformers~\cite{vaswani2017attention_is_all_need} and diffusion models~\cite{ho2020denoising}.

%
While existing methods~\cite{hong2023lrm,wang2023rodin,shi2023zero123++,liu2023syncdreamer,tang2023dreamgaussian} mainly focus on the image-to-3D generation of a single object, this paper explores the more intricate challenge of generating compositional 3D assets:
when presented with an image of an asset containing two compositional objects, our goal is to generate separate 3D representations of each component while ensuring that their relationships are semantically coherent and geometrically and physically plausible.

A simple strategy is to generate the entire assets as a holistic 3D mesh and subsequently use surface segmentation to separate the individual objects, as implemented in Part123~\cite{liu2024part123} and SAMPart3D~\cite{yang2024sampart3d}. However, mesh segmentation usually leads to incomplete surfaces and disregards the relationships among objects.
Alternative methods involve generating each component as an individual object and then combining them into a single model using estimated spatial placement, such as the similarity transformation that includes translation, rotation, and scaling. Examples of this approach can be found in ~\cite{epstein2024disentangled,chen2024comboverse}.
However, they struggle to manage complex spatial relationships that extend beyond simple similarity transformations. 
They fail in situations where non-rigid object deformation is required and often result in shape penetrations.

We observe that physical information, such as supporting relationships, stability, and affordance, can offer valuable clues for generating the shapes of interactive objects. For instance, objects in static scenes should exhibit stability. In a scene depicting ``a frog wearing a sweater", the frog should possess adequate body structure to support the sweater; otherwise, gravity will cause the sweater to fall off.
To this end, we integrate differentiable physical simulations into the process of compositional 3D asset generation.

Specifically, given an input image, we generate consistent multi-view images for both the entire assets, a foreground component, and an inpainted occluded background. The multi-view images are subsequently fitted with 3D Gaussian Splatting~\cite{kerbl3Dgaussians} representations.
Then, to ensure the physical plausibility of the assets, we build upon the Score Distillation Sampling (SDS)~\cite{poole2022dreamfusion} method and introduce a physical simulation-enhanced SDS to further optimize the geometry (i.e., positions of Gaussians) for the objects.
%
%
To ensure visual consistency with the input image, we incorporate image loss, i.e., the difference between the input image and rendered image from the generated object as a complement. 
%

We observe that directly applying the SDS and image loss gradient to update Gaussians' positions results in penetrations and non-physical artifacts.  
Our proposed physical simulation-enhanced SDS delegates updates of Gaussians' positions to the physical simulation instead of the optimizer in the training process. By setting the loss gradient as the initial velocity of the physical simulation,  
the simulator serves as a physics-guided optimizer, which progressively corrects the particle positions by solving the physical system.

%
%
Experiments demonstrate the proposed method can generate physically plausible compositional 3D assets given a single image. Our contributions are as follows:
\begin{itemize}
    \item We design a novel pipeline for image-based compositional 3D asset generation, particularly focusing on interactive objects with strong spatial coupling.
    \item We propose a physical simulation-enhanced Score Distillation Sampling to optimize 3D Gaussians in a physically plausible manner.
    \item We are the first to generate 3D compositional assets without penetration from a single image, facilitating downstream applications.
\end{itemize}

\section{Related Work}

\subsection{Image conditioned 3D Generation}

With the remarkable success of diffusion models~\cite{sohl2015deep} in the 2D domain~\cite{ho2020denoising,rombach2022high}, numerous studies have started investigating how to build 3D generation models.
One approach involves generating 3D assets by distilling knowledge from pre-trained 2D generators~\cite{poole2022dreamfusion,qian2023magic123}.
DreamFusion~\cite{poole2022dreamfusion} proposed Score distillation Sampling (SDS) to optimize a NeRF~\cite{nerf} model with images generated by a 2D generator.
Meanwhile, Magic3D~\cite{qian2023magic123} employed a coarse-to-fine, two-stage strategy to enhance both the speed and quality of the generated models.
The other technical solution involves directly training 3D generators using ground truth 3D data, and training denoising models to produce 3D shapes from image conditions. Notable works include Rodin~\cite{wang2023rodin}, LAS~\cite{zheng2023locally}, 3DShape2VecSet~\cite{zhang20233dshape2vecset}, and CLAY~\cite{zhang2024clay}. 
LRM~\cite{hong2023lrm} reformulated 3D generation as a deterministic 2D-to-3D reconstruction problem.
Synthesizing multi-view consistent images enhances the capabilities of 3D generation or reconstruction, as shown in Zero123++~\cite{shi2023zero123++} and Syncdreamer~\cite{liu2023syncdreamer}.

The aforementioned approach generates 3D data in the form of a single, entangled representation, which is not ideal for numerous downstream applications that require semantically compositional shapes.

\subsection{Compositional 3D Reconstruction and Generation.}
ObjectSDF~\cite{wu2022object} and ObjectSDF++~\cite{wu2023objectsdf++} introduced an object-composition neural implicit representation, which allows separate reconstruction of each piece of furniture within a room, solely based on image inputs.
DELTA~\cite{feng2023learning} presented hybrid explicit-implicit 3D representations, designed for the joint reconstruction of compositional avatars. This includes the integration of components such as the face and body, or hair and clothing, respectively. 
Similar compositional avatars generation with a the SMPL~\cite{SMPL:2015} body pror works can be found in~\cite{hu2023humanliff, dong2024tela, wang2023disentangled, wang2023humancoser}.
AssetField~\cite{xiangli2023assetfield} proposed to learn a set of object-aware ground feature planes to represent the scene and various manipulations could be performed to rearrange the objects.
~\cite{po2023compositional,cohen2023set} jointly optimized multiple NeRFs, each for a distinct object, over semantic parts defined by text prompts and bounding boxes.
~\cite{epstein2024disentangled} and SceneWiz3D~\cite{zhang2023scenewiz3d} eliminated the requirements for user-defined bounding boxes by simultaneously learning the layouts.
Since the text could be problematically complicated when describing complex scenes, GraphDreamer~\cite{gao2024graphdreamer} used scene graphs as input instead.
Frankenstein~\cite{yan2024frankenstein} extended 3D diffusion approach for
building a compositional scene generation tool.

In this paper, we adhere to the SDS-based methodology but incorporate 
physics simulation to address the inherent ambiguity of the 2D-to-3D problem and enhance the physical plausibility of the 3D assets.







\subsection{Physics based 3D generation}
Several attempts have been made to generate physically compatible objects. Aiming to generate physically compatible objects, \cite{chen2024atlas3d} proposed an SDS-based method with rigid-body simulation, which can generate self-supporting objects from text. \cite{guo2024physically} presented a method of generating objects constrained by static equilibrium from a single image. In addition to object geometry generation from texts or images, there are existing works focusing on learning the objects' internal material parameters. In \cite{zhang2025physdreamer}, an approach was proposed to distill dynamic priors from pre-trained video diffusion models by minimizing the discrepancy between physical simulation and diffusion-generated videos. \cite{liu2024physics3d} further utilized a more complex viscoelastic material model to simulate the objects and optimize the physical parameters via SDS. The above methods mainly focus on a single object, approaches are proposed for physically plausible scene reconstruction \cite{ni2024phyrecon}, language-grounded physics-based scene editing\cite{qiu2024feature}. The existing methods above mainly focus on either single-object generation or rigid-body scene generation. In our work, we propose a novel approach for non-rigid compositional asset generation.






\section{Preliminaries}

\subsection{Gaussian Splatting}\label{pre:gaussian}
3D Gaussian Splatting~\cite{kerbl3Dgaussians} (GS) has been proven efficient in 3D reconstruction tasks, due to its high inference speed and rendering quality.

Specifically, 3DGS represents 3D scenes as $N$ Gaussians with attributes $G=\{\mu_i,\Sigma_i,q_i,\alpha_i,c_i\}_{i=1}^N$, where $\mu\in\mathbb{R}^3$ is the center, $\Sigma\in\mathbb{R}^3$ is the scaling factor, $q\in\mathbb{R}^4$ is the rotation quaternion, $\alpha\in\mathbb{R}$ is the opacity value, and $c\in\mathbb{R}^3$ is the color feature.

To render an image, all Gaussians are first projected onto an image plane. Then, volumetric rendering is performed for each pixel in front-to-back depth order to produce the alpha map $A_{rd}$ and color map $I_{rd}$.

We use the following loss function to optimize the Gaussians:
\begin{align}
    \mathcal{L} = (1-\lambda_{1})\mathcal{L}_1(I_{gt},I_{rd}) &+ \lambda_{1}\mathcal{L}_{SSIM}(I_{gt},I_{rd}) \\
    & + \lambda_2 A_{rd} (1 - A_{gt}),
\end{align}
where $I_{gt}$ and $A_{gt}$ are ground-truth image and mask map, $\mathcal{L}_1$ is the L1 loss function, $\mathcal{L}_{SSIM}$ is the structure similarity loss function, and $\lambda_{1,2}$ are the weighting factors.

Given a set of images $\{I_{gt,i}\}_{i=1}^M$, we can train 3DGS:
\begin{align}
    G = GaussianSplatting(\{I_{gt,i}\}_{i=1}^M),
\end{align}
where we eliminate the need for ground-truth mask maps since they can be extracted from images using background removal model~\footnote{https://github.com/OPHoperHPO/image-background-remove-tool}.

\begin{figure*}[htb]
  \centering
  \includegraphics[width=0.9\linewidth]{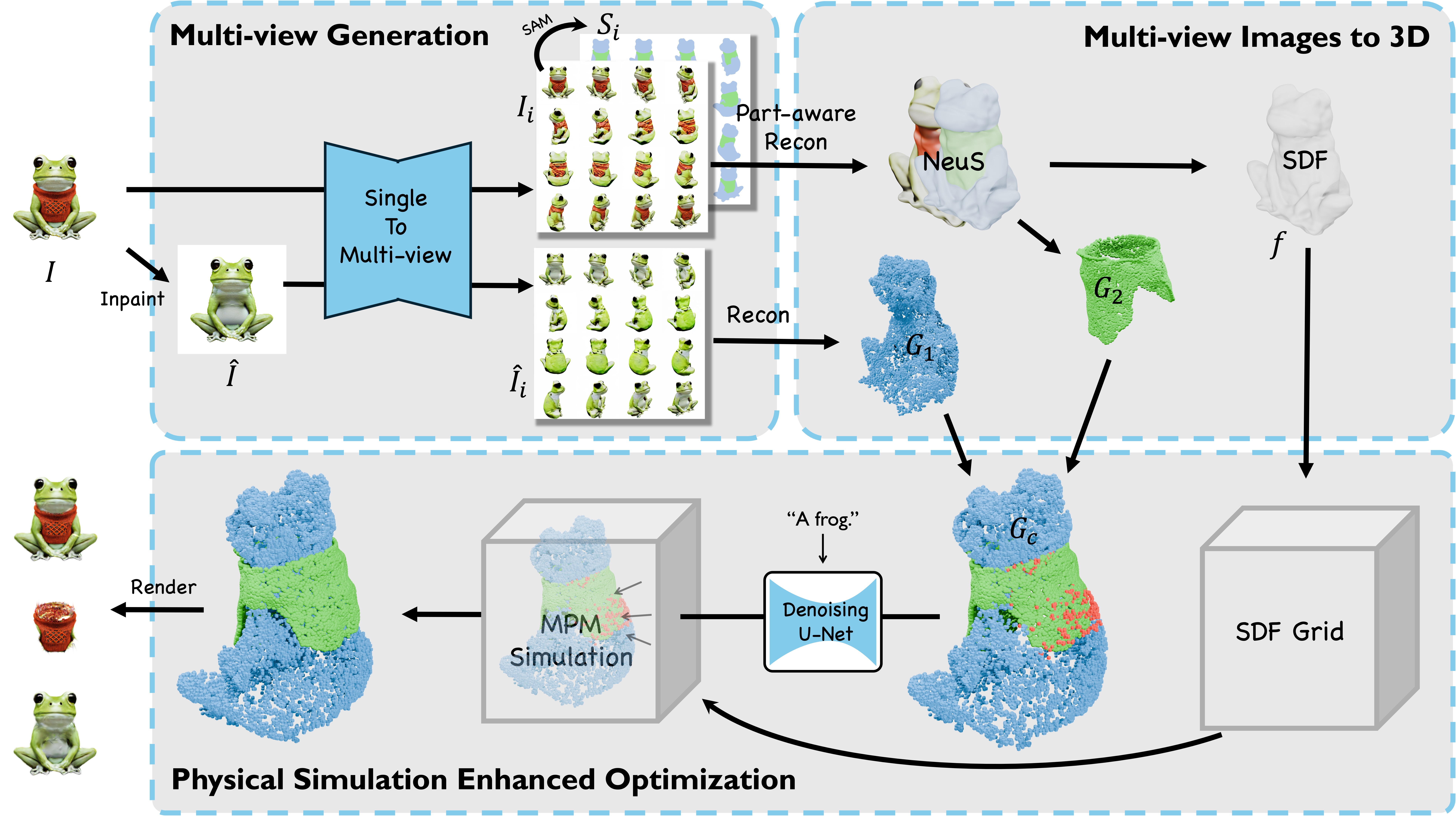}
  \caption{\textbf{The overview of PhyCAGE.} Given an input image, we first generate consistent multi-view images for the components of the assets (see Sec.~\ref{sec:stage1}). 
  Then, we fit multi-view images with 3D Gaussian Splatting representations (see Sec.~\ref{sec:stage2}).
  Finally, we introduce a Physical Simulation-Enhanced SDS to further optimize the positions of the Gaussians (see Sec.~\ref{sec:ps_sds}).
  }
  \label{fig:pipeline}
\end{figure*}

\subsection{Physical Simulation}
\paragraph{Continuum Mechanics.} The motion of material is described by a mapping $\mathbf{x} = \phi(\mathbf{X}, t)$ from rest material space $\mathbf{X}$ to a deformed space $\mathbf{x}$ at time $t$. The Jacobian of the mapping $\mathbf{F}=\frac{\partial\phi}{\partial \mathbf{X}}(\mathbf{X}, t)$, i.e., deformation gradient measures the local rotation and strain \cite{bonet1997nonlinear}. Given the conservation of momentum and conservation of mass, the governing equations for describing the dynamics of an object are as follows:
\begin{equation}
    \rho\frac{D\mathbf{v}}{Dt}=\nabla \cdot \boldsymbol{\sigma} + \mathbf{f}, \quad \frac{D\rho}{Dt} + \rho\nabla\cdot \mathbf{v} = 0,
\end{equation}
where $\mathbf{f}$ denotes an external force, $\boldsymbol{\sigma}$ is the internal stress, the $\mathbf{v}$ and $\rho$ denote the velocity and density respectively.
\paragraph{Material Point Method.}\label{pre:mpm} The Material Point Method (MPM) is a framework for multi-physics simulation. It utilizes the strengths of both Eulerian grids and Lagrangian particles which enables it to simulate phenomena with large deformation, topology changes, and frictional contacts. It is widely adopted for the simulation of a broad range of materials such as elastic objects, snow, sand, and cloth \cite{ram2015material, jiang2015affine, jiang2017anisotropic, hu2018moving, fang2019silly}. Gaussian splatting provides a particle-based explicit 3D representation, which is naturally suitable for serving as the spatial discretization of objects in physical simulation. Following \cite{xie2024physgaussian}, we run MPM on these particles directly. The MPM pipeline consists of three stages in general: particle-to-grid(P2G), grid-operation and grid-to-particle(G2P). In the P2G stage, the MPM  transfers mass and momentum from particles to grids:
\begin{align}
    m_i^n &= \Sigma_p w_{ip}^n m_p \\
    m_i^n \mathbf{v}_i^n &= \Sigma_p w_{ip}^n m_p(\mathbf{v}_p^n + C_p^n(\mathbf{x}_i - \mathbf{x}_p^n)),
\end{align}
where $p$ and $i$ denote the Lagrangian particles and Eulerian grid respectively. The term $w_{ip}^n$ denotes the B-spline basis function defined on the i-th grid, evaluated at the point $\mathbf{x}_p^n$. 
The particles carry properties including position $\mathbf{x}_p^n$, velocity $\mathbf{v}_p^n$, deformation gradient $\mathbf{F_p^n}$, local velocity gradient $\mathcal{C}_p^t$ and mass $m_p$ at timestep $t_n$. The grids are updated after the P2G stage: 
\begin{align}
\mathbf{v}_i^{n+1} = \mathbf{v}_i^{n} - \frac{\Delta t}{m_i}\sum_p\tau^n_p\nabla w_{ip}^n V_p^0 + \Delta tg,
\end{align}
where $g$ denotes the gravity acceleration. The updated velocities are transferred back to the particles as well as updating the positions:
\begin{align}
    \mathbf{v}_p^{t+1} &= \sum_i N (\mathbf{x}_i - \mathbf{x}_p^t)\mathbf{v}_i^t \\
    \mathbf{x}_p^{t+1} &= \mathbf{x}_p^t + \Delta t \mathbf{v}_p^{t+1}.
\end{align}
We utilize the MPM to simulate the interactions of compositional objects in the assets.

\section{Method}

Given an image  $I\in\mathbb{R}^{H\times W}$ of an asset with two compositional objects $\{O_1, O_2\}$ described by text prompts $\tau_1$ and $\tau_2$, we would like to reconstruct a 3D representation of the two objects individually. 
Here we denote $O_1$ and $O_2$ as background and foreground objects respectively. 
We segment the foreground object in image space using Grounded-SAM\cite{kirillov2023segany} to obtain a semantic map. 
The image after segmentation is inpainted to complete the background object. 
For reconstruction, the multi-view images and the inpainted background images are generated using SyncDreamer~\cite{liu2023syncdreamer}. 
We then reconstruct two Gaussian Splatting representations for background and foreground objects, denoted as $G_1$ and $G_2$. A physical simulation-enhanced Score Distillation Sampling (SDS) is then applied to optimize the Gaussians for obtaining a physically plausible representation.

\subsection{Multi-view Generation}\label{sec:stage1}

To reconstruct the object described in the image, we generate the multi-view images from $I$. 
\textbf{First}, we use Grounded-SAM~\cite{kirillov2023segany,liu2023grounding,ren2024grounded} to segment out the masks of both objects:
\begin{align}
    \{M_1,M_2\} = \text{GroundedSAM}(I;\tau_1,\tau_2),
\end{align}
where $M_1,M_2\in\mathbb{R}^{H\times W}$.
\begin{figure}[!b]
  \centering
\includegraphics[width=1.\linewidth]{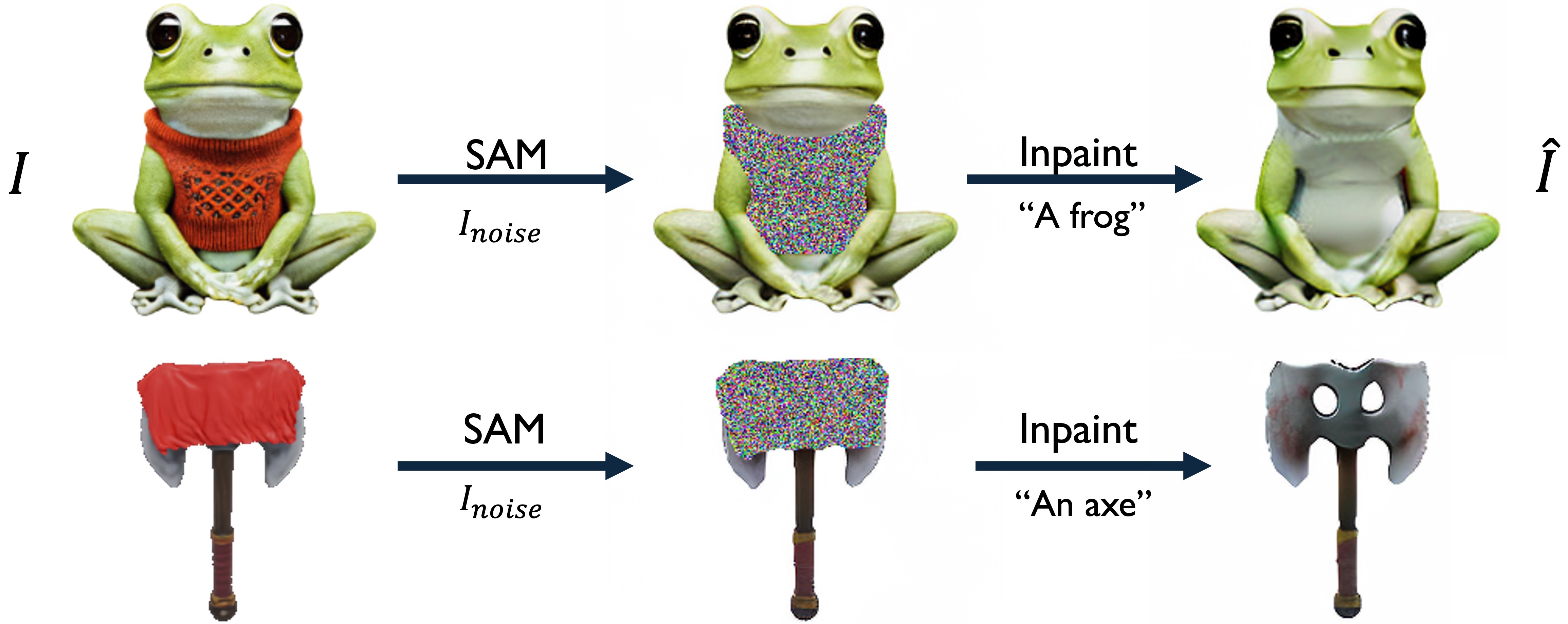}
  \caption{\textbf{Object inpainting with image diffusion models.}}
  \label{fig:inpaint}
\end{figure}
\textbf{Second}, suppose that $O_1$ is occluded by $O_2$, we use inpainting model~\cite{rombach2022high} to complete the image of $O_1$ (see Fig.~\ref{fig:inpaint}):
\begin{align}
    \hat{I} = \text{Inpainting}(I * (\sim M_2) + I_{\text{noise}} * M_2; \tau_1),
\end{align}
where $\hat{I}\in\mathbb{R}^{H\times W}$ is the inpainted image, $I_{\text{noise}}\in\mathbb{R}^{H\times W}$ is an image with random noise sampled from the normal distribution.
\textbf{Third}, we use SyncDreamer~\cite{liu2023syncdreamer} to generate images in 16 different views from $I$ and $\hat{I}$:
\begin{align}
    \{I_i\}_{i=1}^{16} &= \text{SyncDreamer}(I),\\
    \{\hat{I}_i\}_{i=1}^{16} &= \text{SyncDreamer}(\hat{I}),
\end{align}
where $I_i,\hat{I}_i\in\mathbb{R}^{H\times W}$.
\textbf{Furthermore}, we obtain the semantic maps $S_i\in\{-1,1,2\}^{H\times W}$ of each $I_i$ using Grounded-SAM, where -1 refers to the background, 1 refers to $O_1$ and 2 refers to $O_2$.

\subsection{Multi-view Images to 3D}\label{sec:stage2}

We now have 1) the multi-view images and semantic maps $\{I_i, S_i\}_{i=1}^{16}$ of both $O_1$ and $O_2$, and 2) multi-view images $\{\hat{I}_i\}_{i=1}^{16}$ of only $O_1$.
The target is to reconstruct 3D representations from these images and propagate the semantics from 2D images to 3D shapes.
Since GroundedSAM does not guarantee multi-view consistent semantic segmentation, we leverage Part123~\cite{liu2024part123} to integrate the multi-view semantic maps into a 3D consistent one. Specifically,  Part123 optimizes a semantic aware NeuS~\cite{wang2021neus} from $\{I_i, S_i\}_{i=1}^{16}$:
\begin{align}
    \{f, g\} = \text{Part123}(\{I_i,S_i\}_{i=1}^{16}),
\end{align}
where $f:\mathbb{R}^3\mapsto\mathbb{R}$ is the SDF field of both $O_1$ and $O_2$, and $g:\mathbb{R}^3\mapsto\mathbb{R}$ is the 3D semantic field.
By marching cube algorithm, the mesh vertices can be extracted,  denoted as $V=\{v_1,\dots,v_N\}$.
Then $V$ can be split into two groups $V=V_1+V_2$ given the semantics from $g$.
According to our assumption, $V_2$ denotes the mesh vertices of the foreground object $O_2$.
We fit GS for both $O_1$ and $O_2$:
\begin{align}
    G_1 &= \text{GaussianSplatting}(\{\hat{I}_i\}_{i=1}^{16}), \\
    G_2 &= \text{GaussianSplatting}(\{I_i\}_{i=1}^{16};\mu\in{V_2}),
\end{align}
where we keep Gaussian centers of $G_2$ unchanged, i.e., based on the positions of $V_2$, to keep its surface consistent with the extracted SDF. The SDF is utilized as a boundary constraint for the following MPM simulation \cite{fuhrmann2003distance}.

\subsection{Physical Simulation-Enhanced Optimization}
\label{sec:ps_sds}
\paragraph{Score Distillation Sampling Loss.} To ensure the generated Gaussian Splatting $G_1$ is semantically consistent with the description of the inpainted image, we adopt Score Distillation Sampling (SDS) \cite{poole2022dreamfusion} to further optimize its representation. The SDS loss  is defined as:
\begin{align}
    \nabla_\theta \mathcal{L}_{SDS} = \mathbb{E}_{t, \epsilon}\left[ w(t)(\epsilon_{\phi}(I_t^p;y, t) - \epsilon)\frac{\partial I_t^p}{\partial \theta} \right],
    \label{eq:sds_impl}
\end{align}
where $w(t)$ denotes the time-dependent weighting function, $\epsilon_\phi$ represents the pre-trained 2D diffusion model, $I_t^p$ is the predicted image at timestep $t$. Here we reuse the text prompt $y$ for inpainting as the condition for generation.
$\theta$ denotes the parameters of the target Gaussian Splatting representation i.e., $\{\mu,\Sigma,q,\alpha,c\}$ as mentioned in section \ref{pre:gaussian}. Among these parameters, $\mu$ represents the center position for each particle, which is the only key property to take care for ensuring physical plausibility. We freeze opacity $\alpha$ and color $c$ during the optimization to prevent SDS from changing the appearance of the object. Therefore we divided the parameters into three groups $\theta = \{\theta_\mu, \theta_t, \theta_a\}$, where $\theta_t$ denotes the scaling factor and rotation quaternion for the Gaussian particles. $\theta_a$ represents the frozen appearance-related parameters.

\begin{figure}[!ht]

  \centering
  \includegraphics[width=1.0\linewidth]{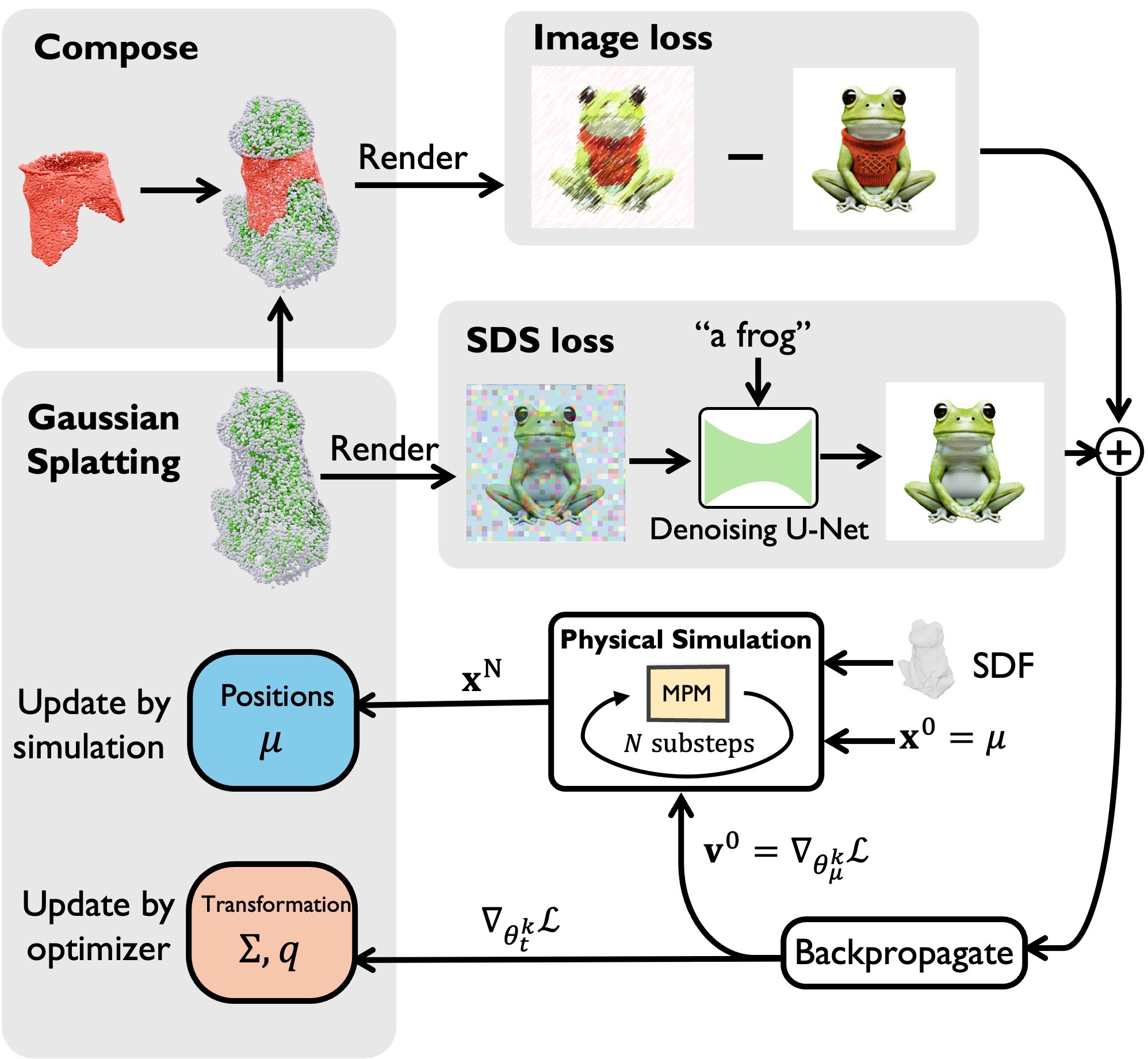}
  \caption{\textbf{The overview of our PSE-SDS}. The gradients come from the SDS and image loss are divided into two streams during the backpropagation. Specifically, $\nabla_{\theta_{\mu}^{k}}\mathcal{L}$ is utilized as the initial velocity of the physical simulation for updating the positions $\mu$ of Gaussians.}
  \label{fig:psel}
\end{figure}

\paragraph{Image Loss.} As we constrain SDS loss for optimizing object geometry only, to ensure visual consistency, we utilize an image loss as a complement to penalize the L1-norm difference between the rendered from the generated composed object $G_c = \{G_1, G_2\}$ and the original input image:
\begin{align}
\mathcal{L}_{Image} = (1-\lambda_1)\mathcal{L}_1(I^c,I) + \lambda_1\mathcal{L}_{SSIM}(I^c, I),
\label{eq:image_loss}
\end{align}
where $I^c$ is the image rendered from the generated composed object, $I$ denotes the original input image, and $\mathcal{L}_{SSIM}$ refers to the structural similarity loss function.

\paragraph{Physical Simulation-Enhanced SDS.}
The final objective is to find parameters $\theta_\mu$ and $\theta_t$, by minimizing
the total loss $\mathcal{L}$:
\begin{align}
\mathcal{L} := \mathcal{L}_{Image}(\theta_{\mu}, \theta_{t}) + \lambda_3\mathcal{L}_{SDS}(\theta_{\mu}, \theta_{t}),
\label{eq:total_loss}
\end{align}
where the $\mathcal{L}_{SDS}$ and $\mathcal{L}_{Image}$ are designed to penalize discrepancy in geometry and visual appearance respectively, between the generated objects and the input image. $\lambda_3$ is the weighting factor.

We observed that directly applying the loss gradient to update particle positions $\mu$ results in penetrations and artifacts as shown in Figure \ref{fig:ab3}. To ensure the physical plausibility, we propose physical simulation-enhanced SDS (shown in Figure \ref{fig:psel}). We delegate the updates of $\mu$ to the physical simulation. Here we use the MLS-MPM \cite{hu2018moving} as the physical simulator. One sub-step of the simulation process can be formalized as follows:
\begin{align}
    \mathbf{x}^{n+1}, \mathbf{v}^{n+1} = \text{MPM}(\mathbf{x}^{n}, \mathbf{v}^{n}, \Delta t, \psi),
\end{align}
where $\mathbf{x}^{n}$ and $\mathbf{v}^{n}$ represent particle position and velocity at timestep $n$, $\psi$ denotes all other properties such as the particle mass, particle volume and materials parameters. Note we omit the subscript $p$ for clarity compared to the notations mentioned in section \ref{pre:mpm}.

As described in algorithm \ref{alg:ps_sds}, given $K$ steps of optimization, we set the $\nabla_{\theta_\mu^k} \mathcal{L}$ i.e., loss gradient with respect to particle position as the initial velocity of particles for the MPM based physical simulation. The MPM outputs the updated $\mu^{k+1}$ after $N$ sub-step simulations.
\begin{algorithm}
\caption{Physical Simulation-Enhanced SDS}
\label{alg:ps_sds}
\begin{algorithmic}[1]
\REQUIRE Given $K$ steps of optimization, $N$ sub-steps MPM simulation, learning rate $\gamma$ 
\FOR{$k=1$ to $K$}
\STATE Compute $\nabla_{\theta^k} \mathcal{L}$ according to Eqn.\ref{eq:total_loss}
\STATE $\nabla_{\theta^k} \mathcal{L} = \{\nabla_{\theta_\mu^k} \mathcal{L}, \nabla_{\theta_t^k} \mathcal{L}\}$
\STATE $\mathbf{x}^{0} = \mu^k, \mathbf{v}^{0} = \nabla_{\theta_\mu^k}\mathcal{L}$
\STATE $\Delta t = \gamma / N$
\FOR{$n = 0$ to $N$}
    \STATE $\mathbf{x}^{n+1}, \mathbf{v}^{n+1} = \text{MPM}(\mathbf{x}^{n}, \mathbf{v}^{n}, \Delta t, \psi)$
\ENDFOR
\STATE $\mu^{k+1} = \mathbf{x}^{N}$
\STATE $\theta_t^{k+1} = \theta_t^{k} - \gamma \nabla_{\theta_t^{k}} \mathcal{L}$
\ENDFOR
\end{algorithmic}
\end{algorithm}

Intuitively, at the first sub-step of the simulation, the MPM advances the particles' positions according to the initial velocity (i.e., loss gradient), which is equivalent to one step of vanilla optimization using gradient descent with a step size $\Delta t$. The following simulation sub-steps are then performed to progressively correct the particles positions by solving the physical system.

\section{Experiment}

\subsection{Implementation Details}

We use Stable-Diffusion-XL-1.0 as an inpainting model with a guidance scale in \{7.5, 8.0, 9.0, 12.5\}. During SDS optimization, we decrease timestep $t$ from 100 to 20. We train NeuS with 1k steps, fit $G_2$ with 30k steps and $G_1$ with 3k steps, and perform the physical simulation-enhanced optimization with 500 steps. We empirically set $\lambda_1=0.2,\lambda_2=1.0,\lambda_3=0.00001$.

\subsection{Evaluation}
We assess the results with the following metrics: 1) Peak Signal-to-Noise Ratio (PSNR), which quantifies the similarity between the rendered image and the input image at the reference view; 2) CLIP score~\cite{radford2021learningtransferablevisualmodels} for various comparisons, including between novel-view images and the input image (CLIP$_{mv}$), between the reference view of $O_1$ and the inpainting prompt (CLIP$_{text}$), between the reference view of $O_1$ and the inpainted image (CLIP$_{ip}$), and between the novel-view images of $O_1$ and the inpainted image (CLIP$_{ip}^{mv}$).


\begin{table}[htb]
    \caption{\textbf{Quantitative comparison with previous work.}}
    \label{tab:sota}
    \centering
    \scalebox{0.9}{
        \begin{tabular}{l|cc}
        \toprule          
        Method & PSNR(dB)$\uparrow$ & CLIP$_{mv}$(\%)$\uparrow$\\
        \midrule
        Part123~\cite{liu2024part123} & 17.52 & 79.60\\
        ComboVerse~\cite{chen2024comboverse} & 16.22 & 85.23\\
        Ours & \textbf{30.70} & \textbf{87.29} \\
        \bottomrule
        \end{tabular}
    }
\end{table}

\begin{figure*}[htb]
  \centering
  \includegraphics[width=1.0\linewidth]{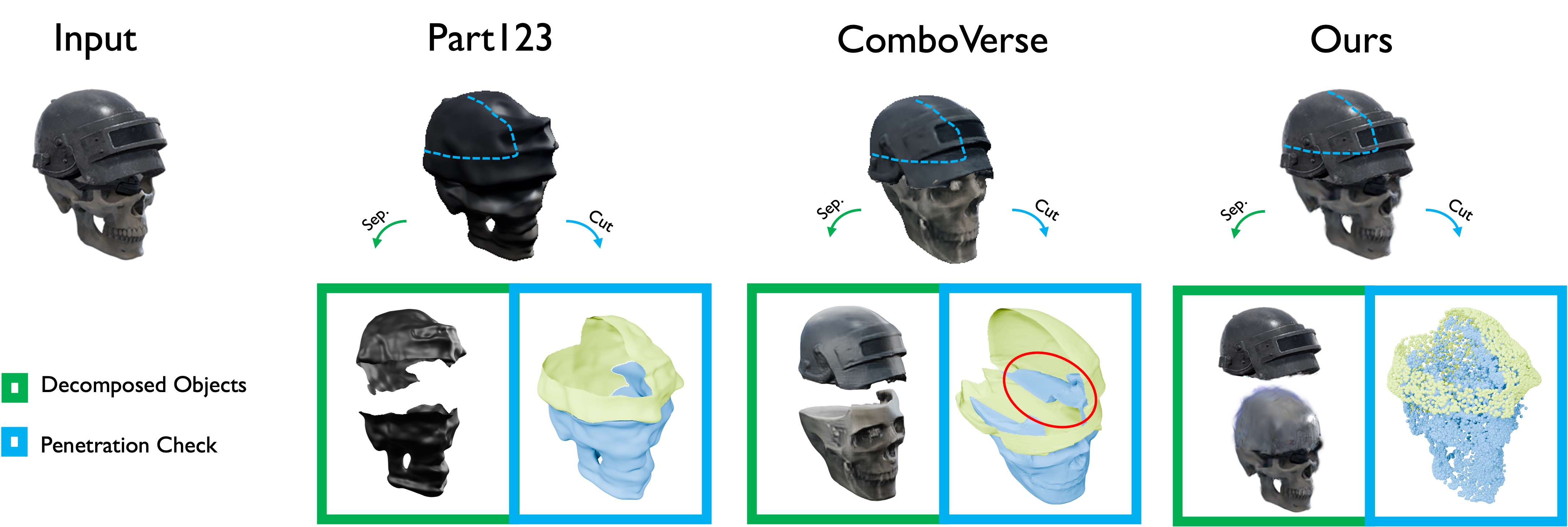}
  \caption{\textbf{Qualitative comparison with previous work.} The green box illustrates the decomposed objects, while the blue box highlights the physical relationships, such as whether the components are in penetration (in red circle). Since we use 3DGS representation, we convert Gaussian centers to point clouds for geometry visualization.}
  \label{fig:sota}
\end{figure*}

\subsection{Comparison with Baseline}

We compare our approach with the following baselines:
1) Part123~\cite{liu2024part123}, which generates a holistic mesh with semantics from a single image;
2) ComboVerse~\cite{chen2024comboverse}, which generates each component in the image separately, and assembles them with estimated similarity transformations.
Fig.~\ref{fig:sota} and Tab.~\ref{tab:sota} shows the qualitative and quantitative results.
Overall, our method produces the most superior 3D compositional assets, taking into account both visual quality and physical plausibility.
Part123 generates the entire assets as a single mesh, leading to incompletely segmented objects.
ComboVerse can not address penetrations between objects.
Our method achieves better consistency with the input image and effectively resolves the penetration problem.

\begin{figure}[htb]
  \centering
  \includegraphics[width=0.95\linewidth]{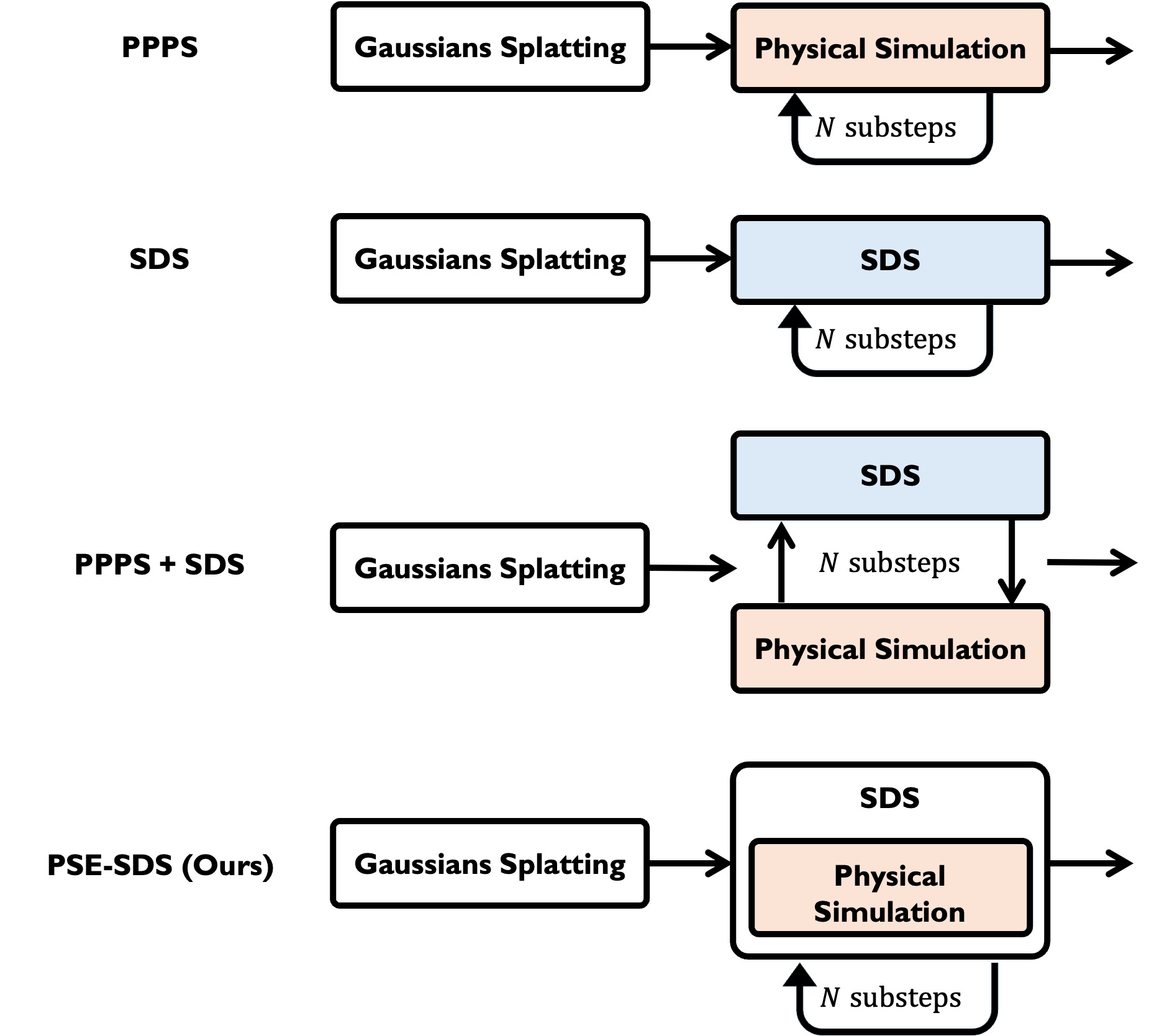}
  \caption{\textbf{Various methods of integrating interactive information through physical simulation}}
  \label{fig:phy_method_compare}
\end{figure}

\subsection{Ablation Study}
We conduct the following ablation study to validate the effectiveness of our Physical Simulation-Enhanced SDS (PSE-SDS):
1) \textbf{PPPS} uses physical simulation as a post-processing procedure after the asset generation. 2) \textbf{SDS} denotes the vanilla SDS optimization that relying solely on visual supervision. 3) \textbf{PPPS + SDS} represents performing simulation and the vanilla SDS alternately. A comparison of how these variants integrate information through physical simulation is shown in Figure \ref{fig:phy_method_compare}.

\begin{table}
    \caption{\textbf{Quantitative results of ablation studies on PSE-SDS.} The unit for PSNR is decibels (dB), and that for CLIP scores is percentage (\%).}
    \label{tab:ab3}
    \centering
    \scalebox{0.75}{
        \begin{tabular}{l|ccccc}
        \toprule          
        Method & PSNR$\uparrow$ &CLIP$_{text}$$\uparrow$ &CLIP$_{ip}$$\uparrow$ &CLIP$_{mv}$$\uparrow$&CLIP$_{ip}^{mv}$$\uparrow$\\
        \midrule

        PPPS & 25.02 & \textbf{29.17} & \textbf{93.66} & 87.36 & 86.86\\
        SDS & 29.13 & 28.35 & 89.76 & 87.90 & 84.83\\
        PPPS + SDS  & 20.74  & 28.17 & 89.46 & 88.06 & 84.69\\
        PSE-SDS (Ours) & \textbf{29.79} & \underline{28.66} & \underline{92.68} & \textbf{88.30} & \textbf{86.93}\\
        \bottomrule
        \end{tabular}
    }
\end{table}

\begin{figure}[!b]
  \centering
  \includegraphics[width=1.\linewidth]{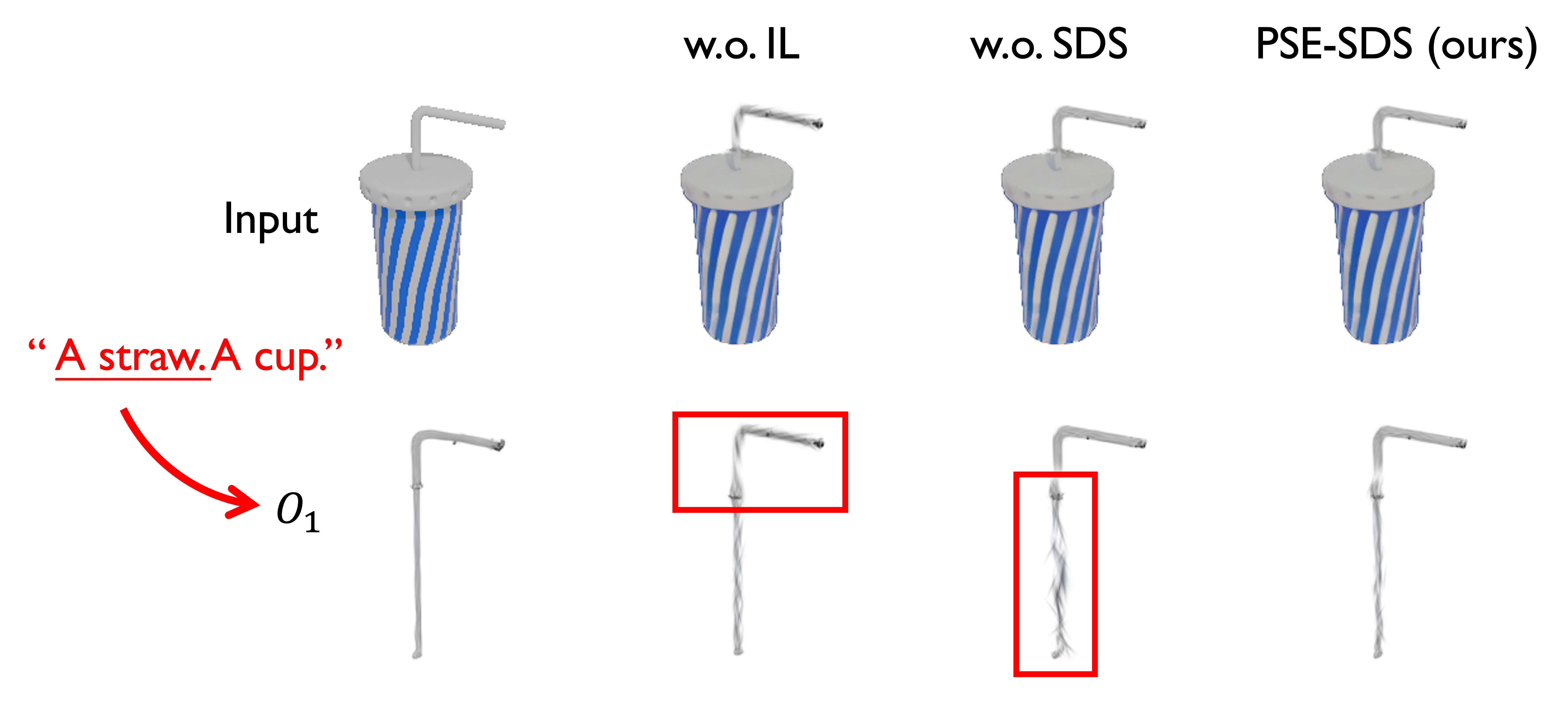}
  \caption{\textbf{Ablation studies on $\mathcal{L}_{Image}$ and $\mathcal{L}_{SDS}$.}}
  \label{fig:abloss}
\end{figure}

\begin{figure*}[htb]
  \centering
  \includegraphics[width=1.0\linewidth]{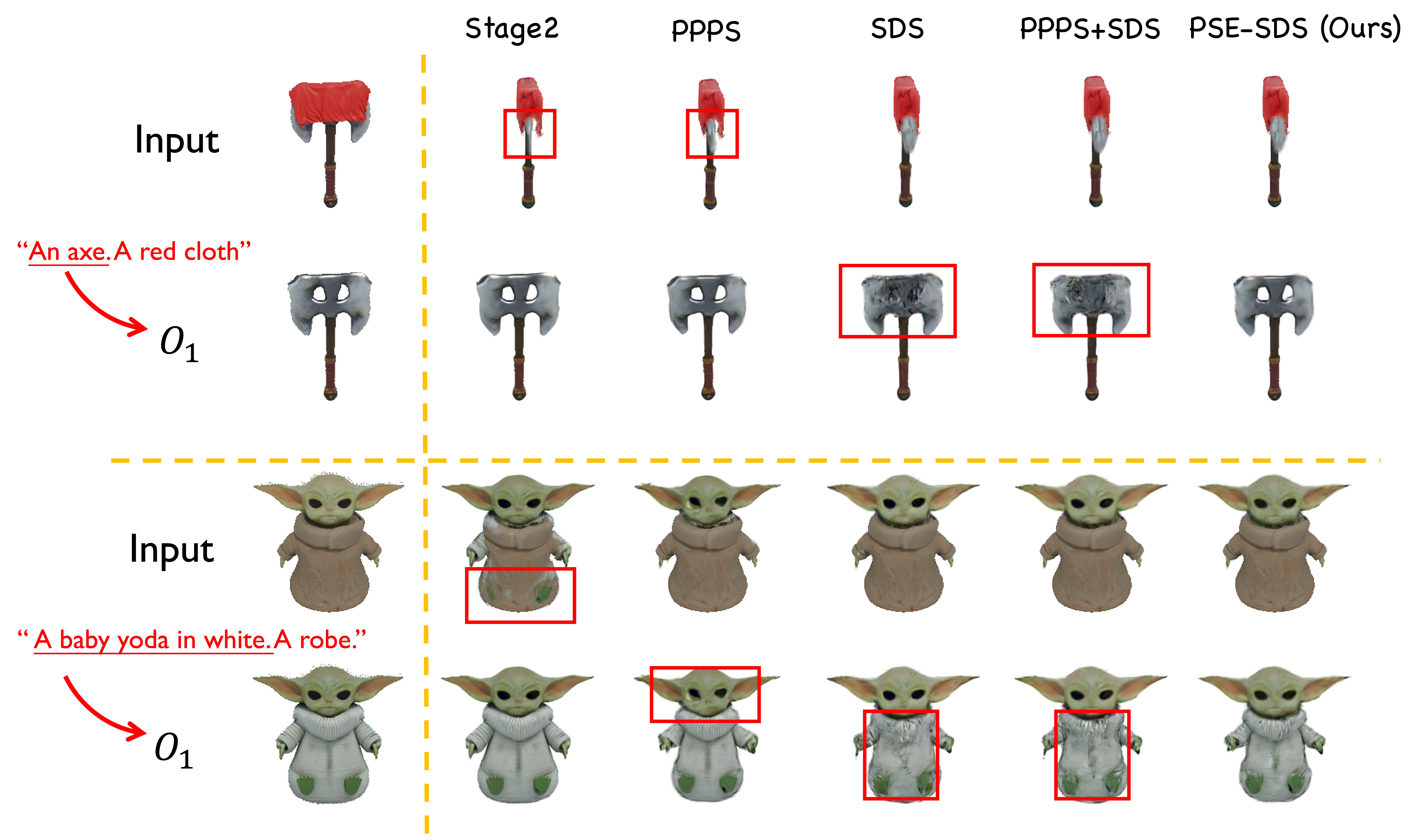}
  \caption{\textbf{Qualitative results of ablation studies on PSE-SDS.}}
  \label{fig:ab3}
\end{figure*}

\paragraph{Effectiveness of Physical Simulation-Enhanced SDS.}

\begin{figure}[!b]
  \centering
  \includegraphics[width=1.\linewidth]{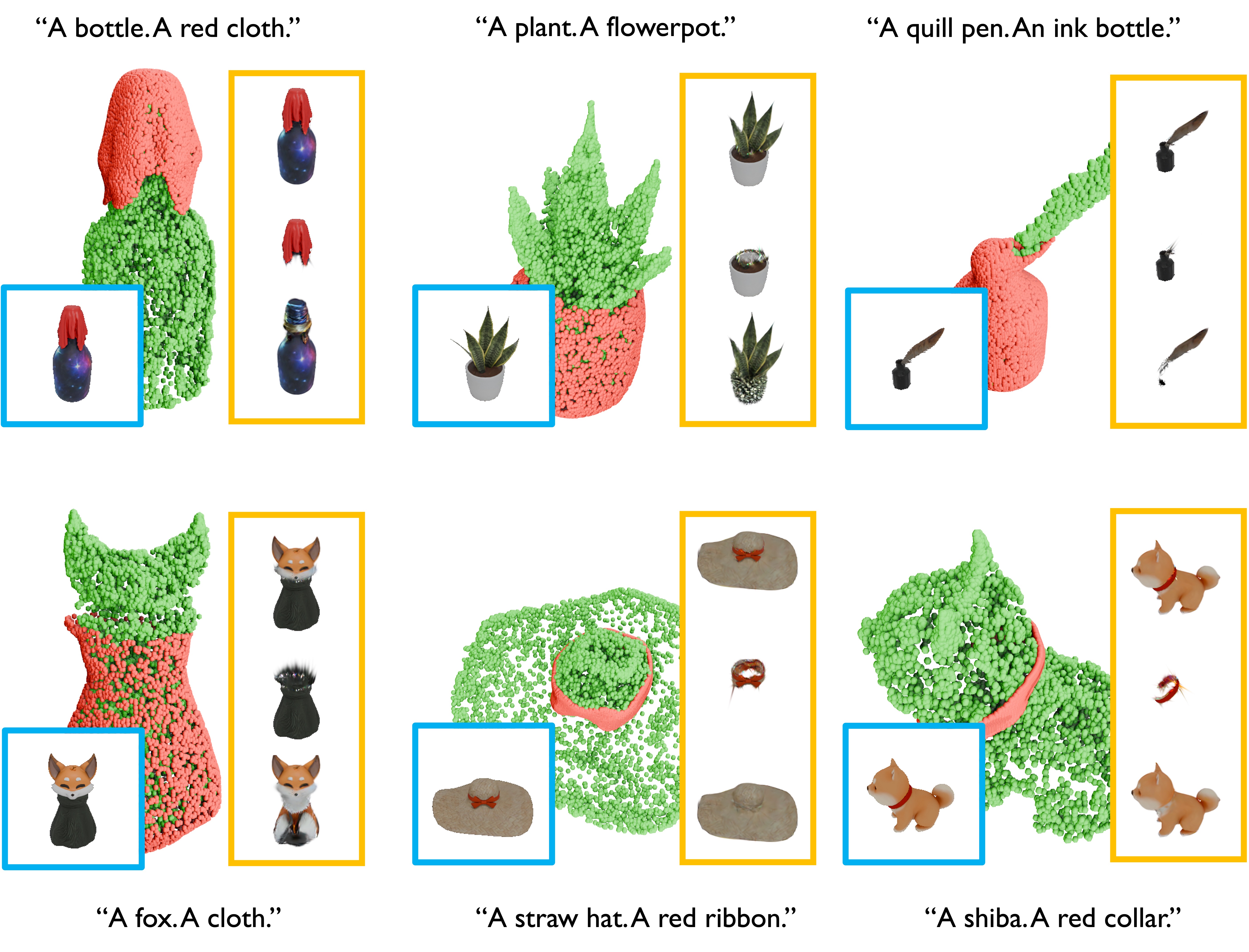}
  \caption{\textbf{More results.} The blue boxes depict the input image, whereas the orange boxes showcase the generated outcomes featuring decomposed objects. The colored point clouds provide visualizations of our generated 3DGS.}
  \label{fig:more_results}
\end{figure}

The output generated in Stage 2 (Sec.~\ref{sec:stage2}) encounters penetration issues (indicated by red boxes in the second column of Fig.~\ref{fig:ab3}), due to the omission of interactive information in the process.
Tab.~\ref{tab:ab3} and Fig.~\ref{fig:ab3} provide both quantitative and qualitative insights into the ablation studies examining various methods of integrating interactive information through physical simulation.
1) \textbf{PPPS} overlooks visual plausibility, since physical simulation treats every point as material without considering semantics.
2) \textbf{SDS} disregards physical plausibility; even though the overall asset aligns well with the input image, individual objects may collapse.
3) \textbf{PPPS+SDS} can still result in object collapse without adequate physical constraints.
4) Our \textbf{PSE-SDS} yields superior outcomes in terms of both visual and physical plausibility.
We present more examples in Figure \ref{fig:more_results} to demonstrate that our method can generate assets with diverse compositional layouts.

\paragraph{Are SDS and Image Loss both necessary?}

We further assess the individual contributions of $\mathcal{L}_{SDS}$ and $\mathcal{L}_{Image}$, respectively (See Fig.~\ref{fig:abloss}).
Excluding $\mathcal{L}_{Image}$ (w.o. IL) leads to the appearance of extraneous object, attributable to the variability inherent in SDS.
Omitting $\mathcal{L}_{SDS}$ (w.o. SDS) results in poor visual plausibility within occluded areas.

\section{Conclusion}
In this paper, we present PhyCAGE, the first approach to generate physically plausible compositional 3D assets from a single Image. 
Our method incorporates a novel Physical Simulation-Enhanced Score Distillation Sampling (PSE-SDS) technique, which leverages a physical simulator as a physics-guided optimizer. This optimizer iteratively corrects the positions of the reconstructed Gaussians to achieve a physically compatible state. 
The experiments demonstrate that PhyCAGE is capable of generating various 3D assets in diverse compositional layouts.
We believe our method represents a significant first step toward physics-aware 3D scene generation.

\smallskip
\noindent
\textbf{Limitations and future work.} Our approach mainly focuses on assets consisting of two objects currently. While it has the potential to be extended to scenarios with more objects by iteratively treating one object as the foreground and the remaining ones as the background during each generation sub-routine.
The quality of the final output depends on the performance of the multi-view generation method. We expect that our approach can be further improved by leveraging more robust reconstruction model in the future.
We aim to further develop our method to generate mesh-based assets, thereby supporting more simulation techniques such as the Finite Element Method (FEM)~\cite{sifakis2012fem} and sophisticated collision handling methods~\cite{Li2020IPC}.
{
    \small
    \bibliographystyle{ieeenat_fullname}
    \bibliography{main}
}


\end{document}